\documentclass[letterpaper,10pt,conference]{ieeeconf}
\IEEEoverridecommandlockouts

\usepackage{cite}
\usepackage{amsmath,amssymb,amsfonts}
\usepackage{algorithmic} 
\usepackage{graphicx}
\usepackage{textcomp}
\usepackage{xcolor}
\usepackage{booktabs}
\usepackage{multirow}
\usepackage{subcaption}
\usepackage{hyperref}

\def\BibTeX{{\rm B\kern-.05em{\sc i\kern-.025em b}\kern-.08em
    T\kern-.1667em\lower.7ex\hbox{E}\kern-.125emX}}

\newcommand{\myindexterms}[1]{%
\vspace{0.5em}
{\small\setlength{\baselineskip}{9pt}%
\noindent\textit{\textbf{Index Terms—}} #1\par}
}

\begin{document}

\title{Verifier-Mediated Multi-LLM Coordination for Multi-Robot Path Planning}

\author{
Vivek Khatana$^{\dagger,\ddagger}$,
Zijian Song$^{\dagger}$,
Naira Hovakimyan$^{\dagger}$,
Petros Voulgaris$^{\ddagger}$\\
{\tt \small $^{\dagger}$\{vkhatana, zijians4, nhovakim\}@illinois.edu} \\
{\tt \small $^{\ddagger}$\{vkhatana, pvoulgaris\}@unr.edu}\\
{\small $^{\dagger}$Department of Mechanical Science and Engineering,
University of Illinois Urbana-Champaign, Urbana, IL, USA} \\
{\small $^{\ddagger}$Department of Mechanical Engineering,
University of Nevada, Reno, NV, USA}
}

\maketitle

\begin{abstract}
Foundation models are increasingly used for high-level multi-robot task planning, yet independently generated symbolic plans can become jointly infeasible after geometric grounding because of shared spatial and temporal constraints. We study execution-aware coordination of large-language-model (LLM) planners for multi-robot navigation. We compare a centralized architecture, in which one LLM jointly reasons over the robot team, with a decentralized multi-LLM architecture in which each robot independently proposes a task plan. A geometric verifier grounds the candidate plans into robot trajectories, detects predicted inter-robot conflicts, and converts conflict information into structured feedback for targeted replanning. This verifier-mediated architecture separates semantic task reasoning from execution-level feasibility while enabling iterative coordination among LLM planners. Experiments in CARLA and on Crazyflie aerial robots evaluate coordination performance as task complexity increases and characterize the trade-offs between centralized reasoning, decentralized verifier-mediated coordination, and physical executability.
\end{abstract}


\myindexterms{\textbf{Large language models, multi-robot systems, multi-agent planning, decentralized coordination, foundation models, robot navigation, path planning}}



\section{Introduction}

Foundation models provide a flexible interface for semantic reasoning, task decomposition, and high-level planning in robotic systems. In multi-robot settings, however, a plan that is individually meaningful for each robot may still be jointly infeasible after geometric grounding. Independently generated routes can intersect in space and time, violate shared workspace constraints, or create execution conflicts that are not apparent at the symbolic planning level. These failures motivate an execution-aware coordination mechanism between foundation-model reasoning and physical robot motion. A key architectural question is whether multi-robot foundation-model planning should be centralized or decentralized. A centralized planner can reason over the complete team state and generate coordinated plans jointly, but its context and decision space grow with team and task complexity. In contrast, decentralized planners provide modularity and local reasoning, but they require an explicit coordination mechanism to resolve conflicts between independently generated plans. Recent multi-robot LLM frameworks coordinate robot teams through centralized or decentralized symbolic reasoning and inter-agent communication~\cite{chen2024centralized,huang2025compositional}. However, such symbolic coordination does not by itself ensure that the resulting plans remain physically feasible after geometric grounding.

\begin{figure}[b!]
    \centering
    \includegraphics[width=0.9\linewidth, trim={75 200 52 120}, clip]{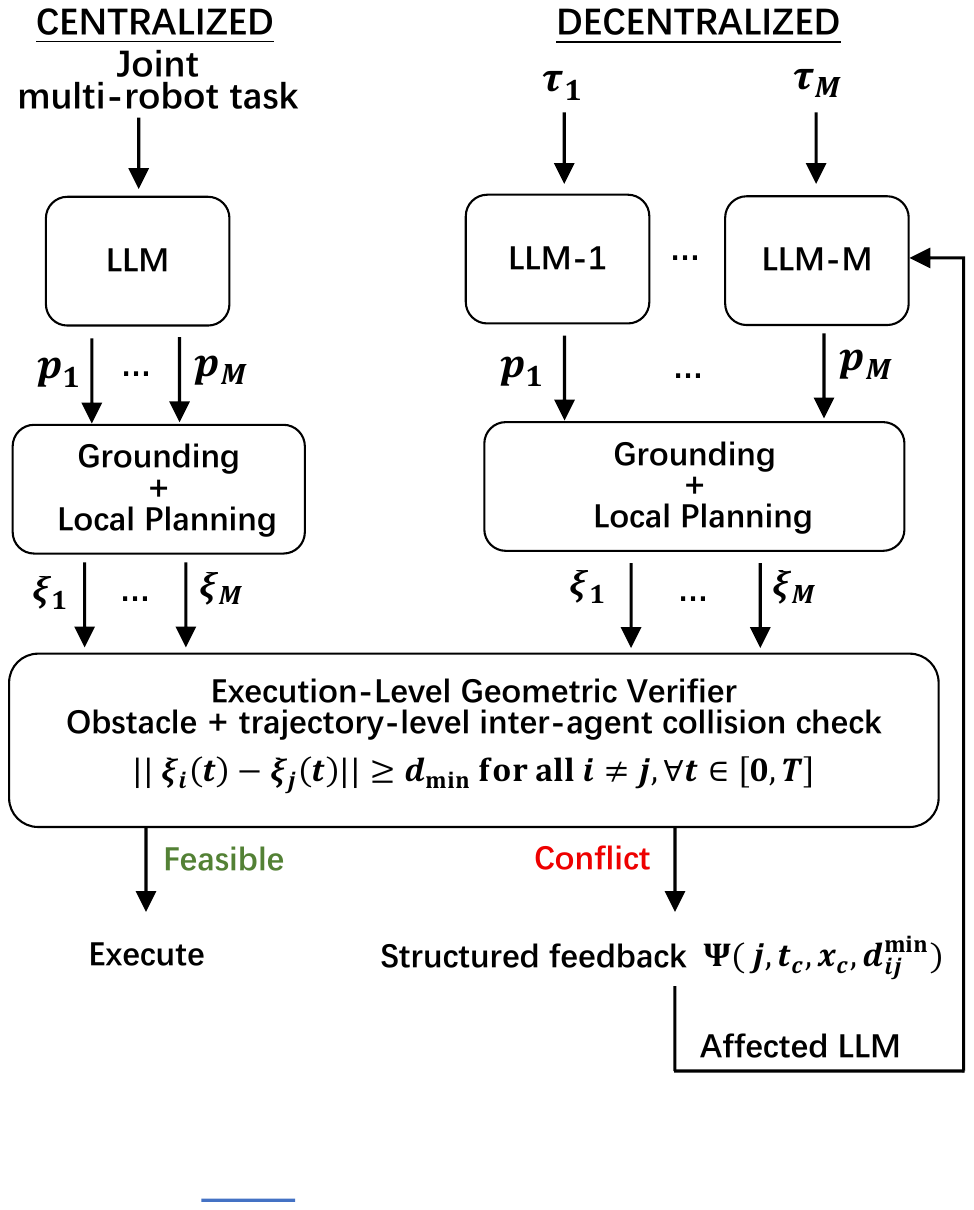}
    \caption{Verifier-mediated multi-robot planning architecture. In the centralized architecture, a single LLM jointly generates symbolic plans for the robot team, whereas in the decentralized architecture each robot is assigned an independent LLM planner. The resulting symbolic plans are grounded into robot trajectories and evaluated by a shared execution-level verifier. Predicted obstacle or inter-agent conflicts are converted into structured feedback and returned to the affected  LLM planner for targeted replanning.}
    \label{fig:architecture}
\end{figure} 

This work studies \textit{verifier-mediated multi-LLM coordination}. Each robot-level LLM proposes a symbolic task plan, which is grounded into an executable trajectory by a geometric planning layer. A shared execution-level verifier evaluates the grounded trajectories for obstacle and inter-robot conflicts. When a conflict is predicted, the verifier converts the failure into \textit{structured feedback}---including the affected agents, conflict location, and conflict time---and returns this information to the appropriate LLM planner for targeted replanning. This creates a closed coordination loop between foundation-model reasoning and robot execution without requiring unrestricted natural-language negotiation. The framework is illustrated in Fig.~\ref{fig:architecture}. 

\noindent The main contributions of this work are:
\begin{itemize}
    \item[i)] We formulate centralized and decentralized LLM-based architectures for multi-robot task planning under shared execution constraints.
    \item[ii)] We introduce a verifier-mediated coordination mechanism that converts predicted geometric conflicts into structured feedback for targeted multi-LLM replanning.
    \item[iii)] We empirically study coordination and task completion performance across increasingly complex multi-robot tasks in CARLA~\cite{Dosovitskiy17} and demonstrate transfer of the planning--verification--execution framework to Crazyflie~\cite{bitcraze_crazyflie20} aerial robots.
\end{itemize}

\textbf{Relation to prior work.} Recent work has explored centralized and decentralized LLM-based multi-robot  symbolic reasoning and coordination~\cite{chen2024centralized,huang2025compositional,li2026survey}, as well as feedback and constraint-guided refinement of LLM-enabled robotic behavior~\cite{madaan2023selfrefine,huang2022innermonologue,ravichandran2026safety}. Our focus is different: we use an external execution-level verifier to evaluate grounded multi-robot trajectories, detect physical conflicts, and return structured agent-specific geometric feedback for targeted multi-LLM replanning.

\section{Verifier-Mediated Multi-Robot Planning}\label{sec:LLM_mediation}

\subsection{Multi-Robot Task and Path Representation}
We represent the robot environment by a graph $\mathcal{G}=(\mathcal{V},\mathcal{E})$, where nodes denote admissible symbolic states and edges denote valid transitions. Robot $i\in\{1,\ldots,M\}$ is assigned a task $\tau_i := \{s_i,g_i,\mathcal{W}_i,\mathcal{F}_i,\mathcal{C}_i\}$ containing its start $s_i$, goal $g_i$, required targets $\mathcal{W}_i$, and task-specific constraints $(\mathcal{F}_i,\mathcal{C}_i)$. The LLM generates a symbolic plan
\begin{equation}
    p_i = \mathcal{P}_{\mathrm{LLM},i}(\tau_i,\mathcal{G},I_i),
    \label{eq:llm_plan}
\end{equation}
where $I_i$ denotes the planner context, including prior execution-level feedback when available. Each symbolic plan is grounded and passed to a local geometric planner,
\begin{equation}
    \xi_i = \Pi_{\mathrm{loc}}\!\left(\Gamma(p_i)\right),
    \label{eq:grounding}
\end{equation}
where $\Gamma:\mathcal{V}\rightarrow\mathcal{X}$ maps symbolic states to geometric anchors in the robot's state space $\mathcal{X}$ and $\Pi_{\mathrm{loc}}$ is a local geometric planner used to connect grounded waypoints, such as an A$^\star$-based planner~\cite{carlagrp}. The resulting route is subsequently associated with the robot execution trajectory $\xi_i$. 

\subsection{Centralized and Decentralized LLM Architectures}
In the \emph{centralized} architecture, a single LLM receives the joint team task and generates all robot plans simultaneously,
\begin{equation}
    \{p_1,\ldots,p_M\}
    = \mathcal{P}^{c}_{\mathrm{LLM}}(\tau_1,\ldots,\tau_M,\mathcal{G}, I).
    \label{eq:centralized}
\end{equation}
This exposes the planner to the global coordination problem but increases the size of the joint planning context. In the \emph{decentralized} architecture, robot $i$ has its own LLM planner as in \eqref{eq:llm_plan}. The planners need not directly negotiate in unrestricted natural language. Instead, their candidate trajectories are coupled through a shared execution-level verifier.

\subsection{Execution-Level Conflict Verification and Feedback}
For every robot pair $(i,j)$, the verifier evaluates the trajectories and requires
\begin{equation}
    \|\xi_i(t)-\xi_j(t)\| \geq d_{\min},
    \quad \forall i\neq j,\; \forall t\in[0,T],
    \label{eq:separation}
\end{equation}
where $d_{\min}>0$ is a prescribed separation threshold. The verifier can additionally check robot-obstacle constraints using the same grounded trajectories. If condition~\eqref{eq:separation} is violated, the verifier identifies a predicted   and constructs a structured feedback object $\psi$
\begin{equation}
    \begin{aligned}
        \psi_i^{(r)} &= \Psi \left(j,t_c,x_c,d_{ij}^{\min}\right),
        \\
        p_i^{(r+1)} &= \mathcal{P}_{\mathrm{LLM},i}(\tau_i,\mathcal{G},I_i\cup\psi_i^{(r)}),
        \label{eq:feedback}
    \end{aligned}
\end{equation}
where $t_c$ and $x_c$ denote the predicted conflict time and location, $d_{ij}^{\min}$ is the minimum predicted separation between robots $i$ and $j$, and the superscript $r$ denotes the planning iteration $r$. Replanning is targeted to the affected agent while feasible plans from other agents may be retained. The process terminates when all candidate trajectories satisfy the execution-level constraints.

\begin{figure}[t!]
    \centering
    \includegraphics[width=0.8\linewidth, trim={10 10 8 10}, clip]{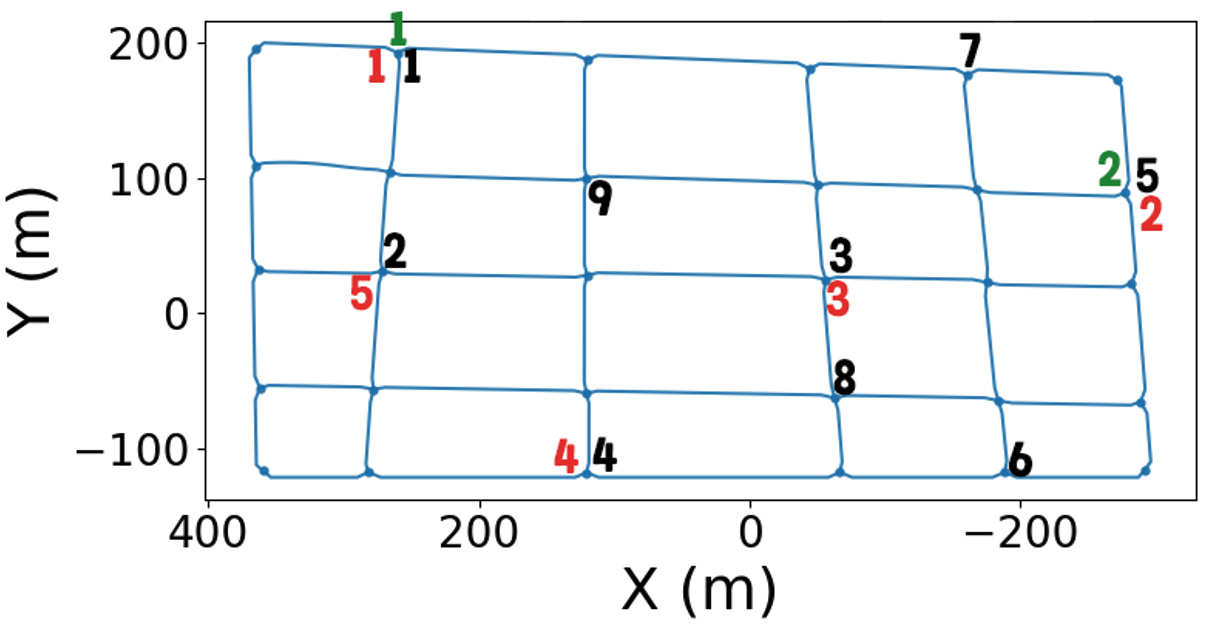}
    \caption{\small CARLA multi-agent, multi-target navigation environment. The two agents navigate the shared road-network graph while visiting target sets of increasing complexity, with $|\mathcal{W}|\in\{2,5,9\}$ corresponding to the green, red, and black task configurations, respectively.}
    \label{fig:multi_target_complexity}
\end{figure}

\begin{figure*}[!t]
\centering
\begin{subfigure}[t]{0.328\textwidth}
    \centering
        \includegraphics[width=\linewidth,trim={10 0 5 0},clip]{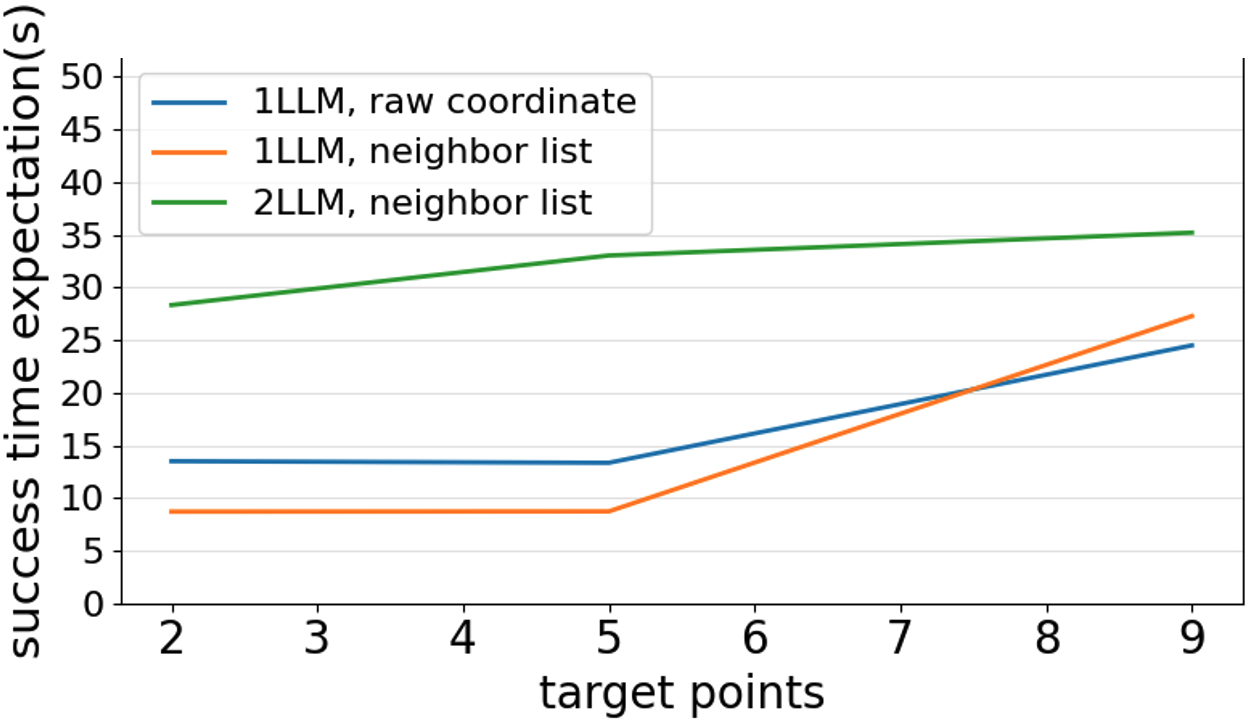}
        \caption{Mean Success Time}
        \label{fig:success_time_comparison}
\end{subfigure}
\begin{subfigure}[t]{0.328\textwidth}
    \centering
        \includegraphics[width=\linewidth, trim={10 0 5 0},clip]{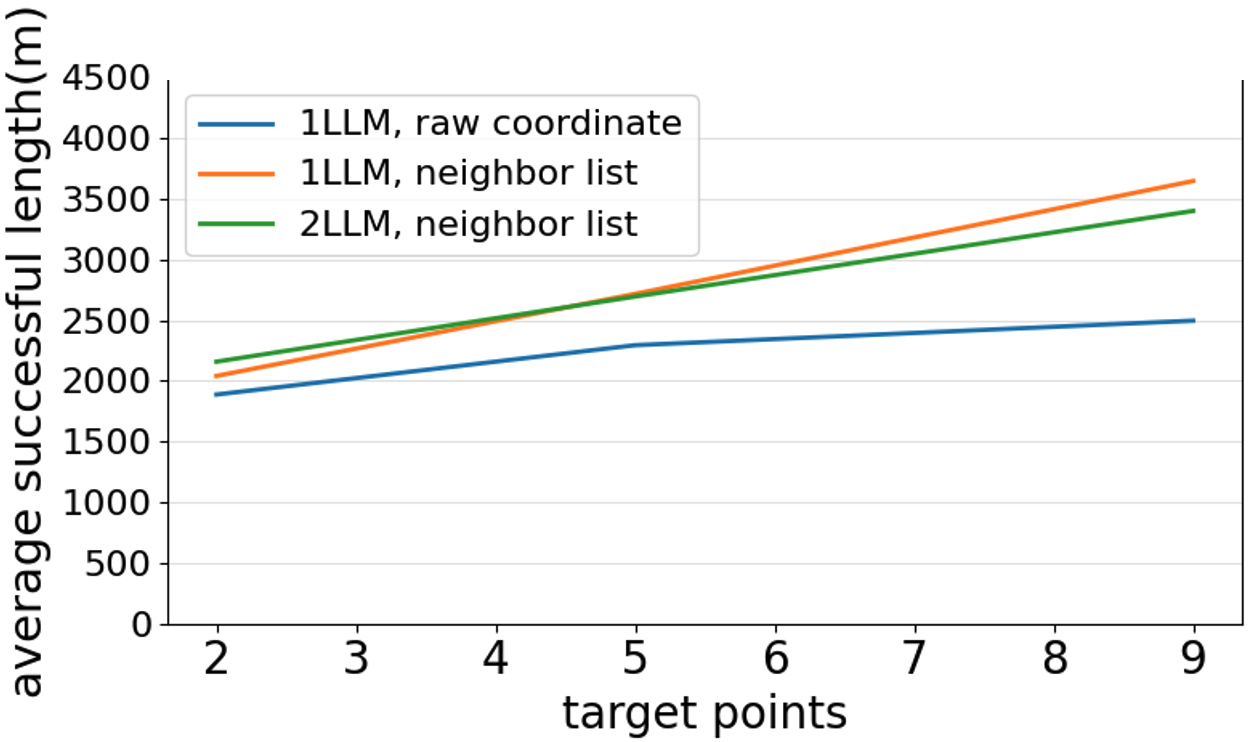}
        \caption{Mean Successful Path Length}
        \label{fig:success_length_comparison}
\end{subfigure}
\begin{subfigure}[t]{0.328\textwidth}
    \centering
        \includegraphics[width=\linewidth, trim={10 0 5 0},clip]{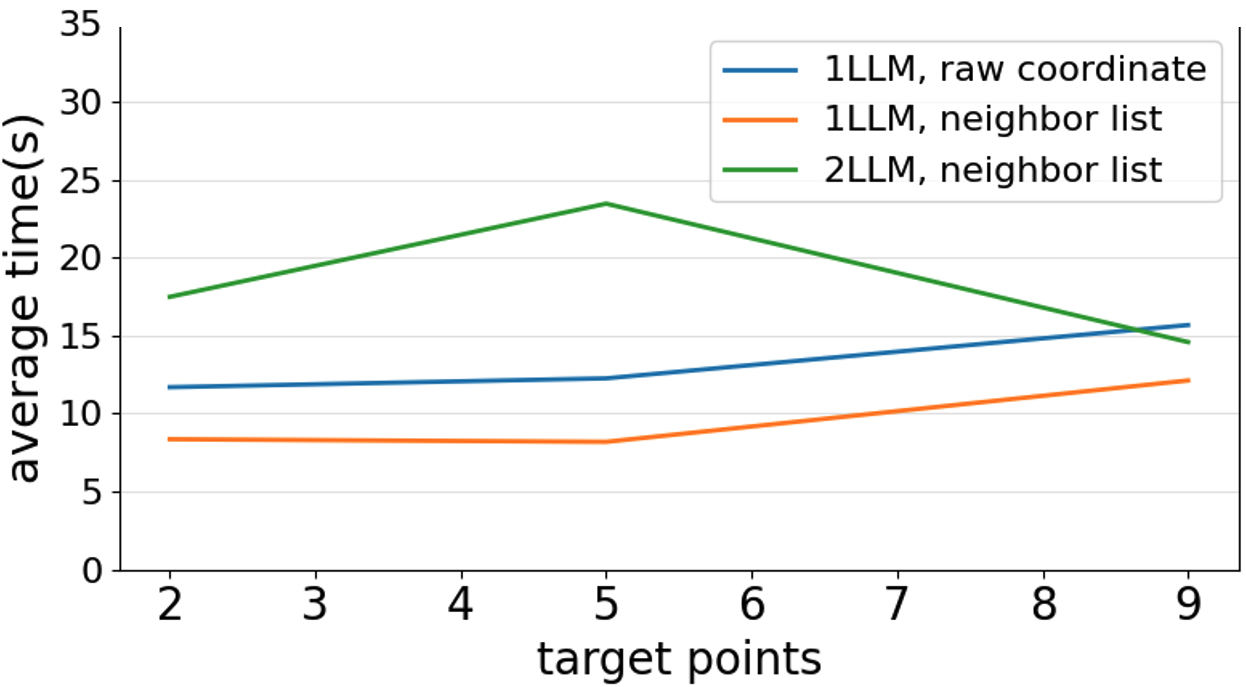}
        \caption{Mean Planning Time Across All Trials}
        \label{fig:raw_time_comparison}
\end{subfigure}
\caption{\small Performance on two-agent multi-target navigation in CARLA as the number of target points increases from 2 to 5 to 9. Results compare centralized single-LLM planning with coordinate and neighbor-list representations against decentralized two-LLM planning with the neighbor-list representation. Metrics are (a) mean planning time of successful trials, (b) mean successful path length, and (c) mean planning time over all trials.}
\label{fig:comparison}
\end{figure*}

\section{Multi-Robot Evaluation}

\subsection{CARLA Experiments}
We first evaluate the proposed centralized and decentralized multi-LLM planning architectures in CARLA~\cite{Dosovitskiy17} using a structured road-network graph (shown in Fig.~\ref{fig:multi_target_complexity}). We evaluate two environment representations: raw coordinate descriptions and a neighbor-list representation. In the raw-coordinate representation, the environment is provided to the LLM through the serialized geometric positions of the graph nodes. In the neighbor-list representation, each node $v_i\in\mathcal{V}$ is instead described by its adjacent nodes, $\mathcal{N}(v_i)=\{v_j:(v_i,v_j)\in\mathcal{E}\}$, thereby explicitly encoding the admissible graph connectivity. We consider a two-agent multi-target navigation problem. Each agent begins at an assigned start node, visits all designated target points in any order, and returns to its start node. The agents share the road network and must avoid inter-agent conflicts during execution. To evaluate increasing task complexity, we consider target sets of size $|\mathcal{W}|\in\{2,5,9\}$ nodes.
A trial is declared successful only if both agents visit all assigned target points and remain collision-free. Our experiments use an inter-agent separation threshold of $7$ m; a trial is declared unsuccessful if the distance between the agents falls below this threshold at any time.

We compare centralized and decentralized planning under the same task conditions. In the centralized architecture, a single LLM jointly generates plans for both agents. In the decentralized architecture, each robot is assigned an independent LLM planner that reasons over its own task and tentative trajectory. Coordination is achieved through iterative verifier-mediated replanning as described in Section~\ref{sec:LLM_mediation}. At the end of each planning iteration, the agents exchange their tentative paths, which are evaluated by a prediction-based collision checker using a
time-indexed distance criterion (cf.~\eqref{eq:separation}). If an inter-agent conflict is detected, the predicted conflict location and time index are communicated to the corresponding LLM, which replans its trajectory while the other agent retains its current plan (cf.~\eqref{eq:feedback}). This process is repeated until no predicted inter-agent conflict is detected.

Performance is evaluated using three primary metrics: (i) expected success time, defined as the average planning time over successful trials; (ii) expected successful path length, defined as the average path length over successful trials; and (iii) expected planning time, defined as the average planning time over all trials. For decentralized planning, the reported planning time is averaged across the two LLM planners. 

Fig.~\ref{fig:comparison} highlights the trade-off between centralized and decentralized LLM coordination as task complexity increases. For smaller target sets, centralized planning incurs lower planning overhead, whereas the performance gap narrows for larger target sets as decentralized planning benefits from its modular structure. The results also show that the environment representation matters: neighbor-list inputs consistently produce graph-valid routes with path lengths comparable to centralized planning, while raw-coordinate inputs provide greater geometric flexibility but are more susceptible to invalid waypoint generation.

\begin{figure*}[!t]
\centering
\begin{subfigure}[t]{0.32\textwidth}
    \centering
        \includegraphics[width=\linewidth,trim={0 100 15 20},clip]{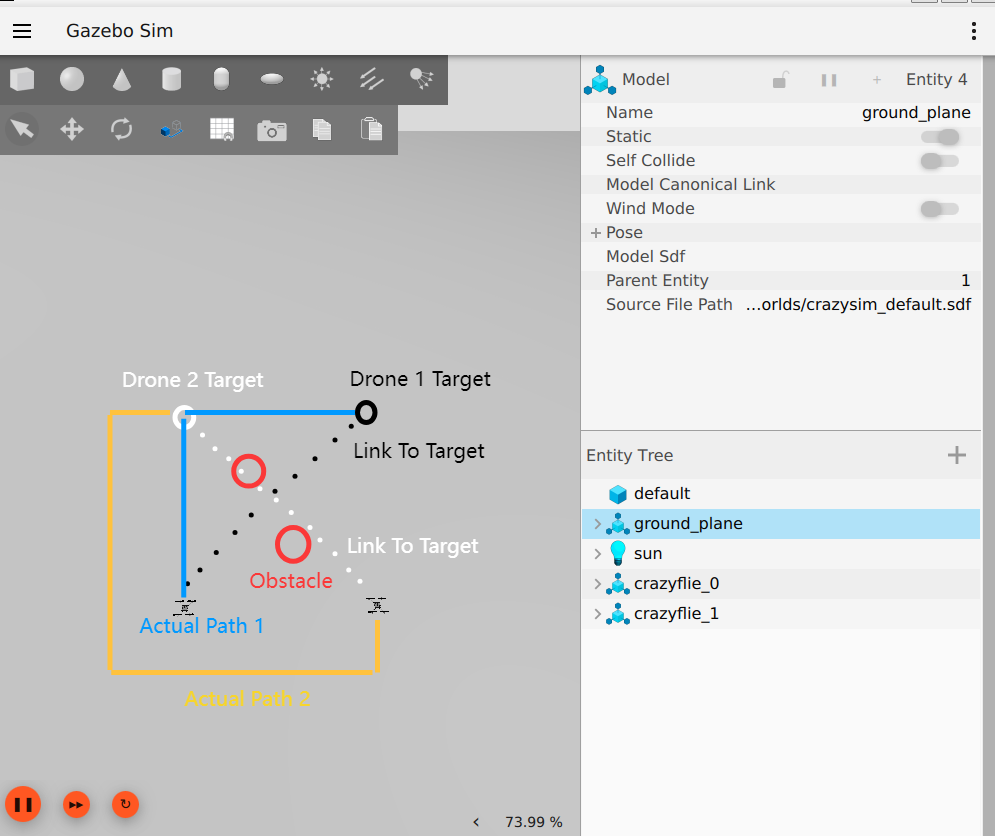}
        \caption{Two-Crazyflie simulation}
        \label{fig:two_crazyflie_sim}
\end{subfigure}
\hfill
\begin{subfigure}[t]{0.65\textwidth}
    \centering
    \includegraphics[width=0.495\linewidth, trim={0 8 0 8}, clip]{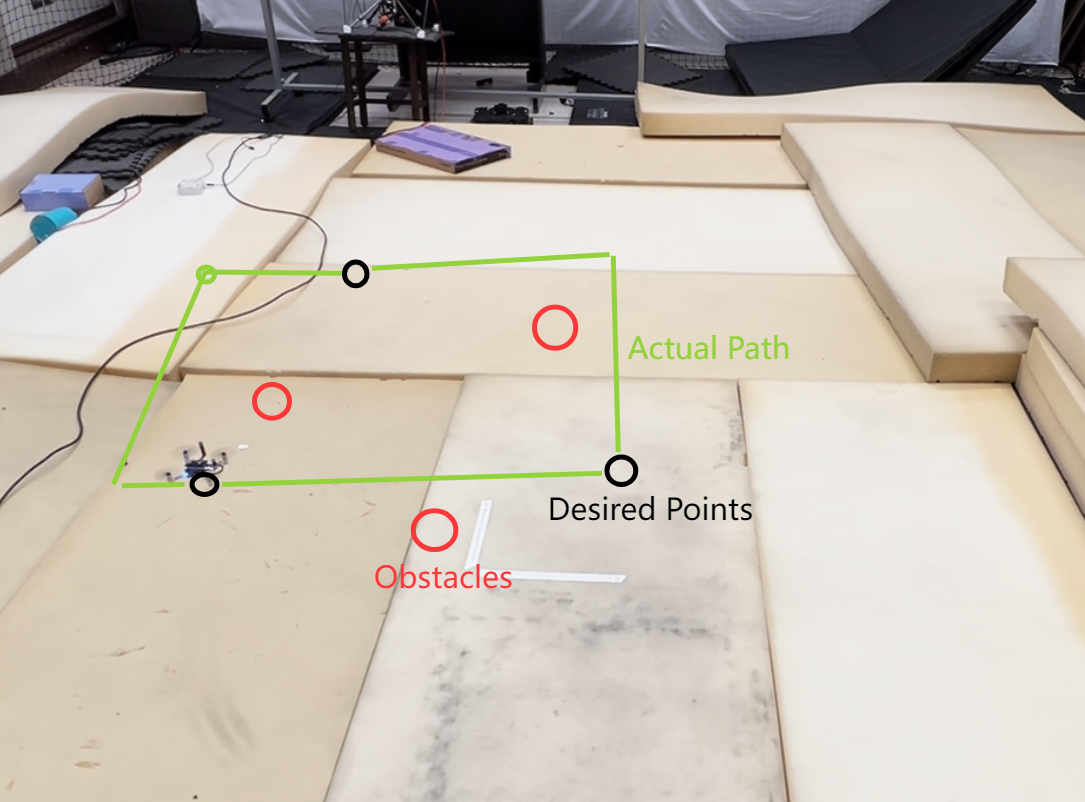}
    \hfill
    \includegraphics[width=0.495\linewidth, trim={0 0 0 0}, clip]{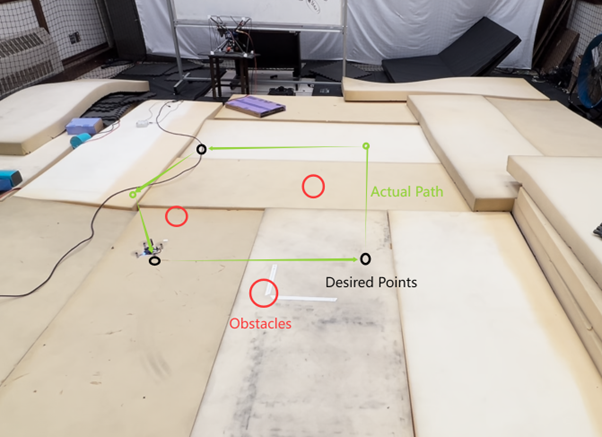}
    \caption{Physical Crazyflie execution.}
    \label{fig:one_crazyflie_hardware}
\end{subfigure}
\caption{\small Cross-platform evaluation on Crazyflie aerial robots.
(a) Two-Crazyflie coordinated navigation in simulation with obstacle
and predicted inter-agent conflict checking (demo video: \url{https://tinyurl.com/cn3anffv}). (b) Physical Crazyflie
navigation demonstrating transfer of the symbolic-to-geometric
planning and feasibility interface to hardware (demo videos: \url{https://tinyurl.com/476s68br}). Multi-robot coordination is evaluated in simulation, while the hardware
experiment evaluates the planning pipeline under tracking, localization, actuation, and physical execution effects.}
\label{fig:crazyflie}
\end{figure*}

\subsection{Insights from CARLA Experiments}

The CARLA experiments reveal several system-level trade-offs in
foundation-model-based multi-robot planning. First, centralized planning is computationally efficient for the two-agent tasks considered here because a
single LLM reasons jointly over the complete multi-agent state and can account
for inter-agent interactions during plan generation. In contrast, decentralized planning introduces additional communication and replanning overhead through verifier-mediated coordination. As task complexity increases, however, the performance gap between the two architectures becomes smaller, suggesting that the
additional coordination overhead of decentralized planning is partially offset
by its modular planning structure.

Second, the environment representation provided to the LLM has a significant
effect on planning reliability. The neighbor-list representation consistently
produces graph-valid feasible routes with similar path lengths across the two
planning architectures, whereas coordinate-based descriptions occasionally
produce shorter trajectories but are more susceptible to invalid or out-of-map
waypoints. This illustrates a trade-off between geometric flexibility and
symbolic validity.

Finally, collision-free task completion became more difficult as the number of target points increased, consistent with longer trajectories, larger waypoint-ordering spaces, and more frequent inter-agent conflicts. These factors also increase the difficulty of resolving conflicts through iterative replanning, highlighting the importance of explicit execution-level feasibility checking. 

\subsection{Crazyflie Cross-Platform Evaluation}

We next evaluate whether the planning--verification--execution interface transfers from ground vehicles to aerial robots through both Crazyflie~\cite{bitcraze_crazyflie20} simulation and physical hardware experiments. In simulation, two quadrotors are assigned distinct navigation tasks, and the same centralized/decentralized planning architectures are used
to generate trajectories satisfying the verifier's predicted inter-agent separation criterion. The geometric verifier evaluates synchronized inter-agent distances and triggers replanning when a predicted conflict is detected. 

On physical hardware, we deploy the same symbolic-to-geometric 
planning pipeline for obstacle-aware navigation of a single Crazyflie. The LLM generates sparse waypoint sequences, which are grounded into executable trajectories and evaluated for feasibility before being passed to the Crazyflie low-level controller. Unlike in simulation, the executed trajectory naturally deviates
from the ideal waypoint connections due to tracking errors, actuator dynamics, localization uncertainty, and external disturbances. The hardware experiments therefore evaluate the planning--grounding--feasibility interface under
execution-level effects, while multi-robot coordination itself is evaluated in the two-Crazyflie simulation experiments. Fig.~\ref{fig:crazyflie} shows some snapshots from the Crazyfile simulation and hardware experiments illustrating that the same planning–verification–execution interface supports both two-robot coordination in simulation and physically executed obstacle-aware navigation on Crazyflie hardware.

\subsection{Insights from Crazyflie Experiments}
The Crazyflie experiments provide complementary evidence of both multi-robot coordination and cross-platform transfer. In simulation, the verifier-mediated architecture enables two independently tasked quadrotors to coordinate through trajectory-level conflict checking and replanning, demonstrating that the proposed interface is not specific to the CARLA road-network setting. The physical experiments further show that the same symbolic-to-geometric planning and feasibility pipeline can be deployed on real aerial hardware despite tracking errors, actuator dynamics, localization uncertainty, and external disturbances. Together, these results highlight the modularity of separating foundation-model reasoning from execution-level feasibility evaluation: the high-level planning architecture can be retained across robot embodiments while the geometric grounding and execution layers are adapted to the underlying platform. In the present implementation, the verifier operates as an execution-level feasibility checker for predicted trajectory conflicts; extending this interface with uncertainty-aware or formally certified safety mechanisms provides a natural direction for future work.

\section{Conclusion}
We presented a verifier-mediated architecture for coordinating foundation-model planners in multi-robot navigation. The approach separates high-level semantic reasoning from execution-level feasibility: centralized or decentralized LLM planners propose robot-level task plans, while a geometric verifier evaluates the grounded trajectories and returns structured conflict information for targeted replanning. This formulation provides a concrete interface between decentralized foundation-model reasoning and execution-aware multi-robot coordination. CARLA experiments study coordination performance under increasing task complexity, while Crazyflie experiments demonstrate transfer of the planning--verification--execution interface to aerial robot navigation. 

Ongoing work includes extending the framework to larger robot teams, richer communication policies, uncertainty-aware verification, and multi-robot hardware deployment.

\section*{{Acknowledgment}}
The authors used ChatGPT (OpenAI) to edit the manuscript. The authors independently conceived all research questions, designed the architectures, experimental methods, and conducted all validations, interpretations, and scientific conclusions.

\bibliographystyle{IEEEtran}
\bibliography{references}  

\end{document}